\documentclass{ws-jmrr}
\usepackage{amsmath,amssymb}
\usepackage{graphicx}
\usepackage{array}
\usepackage{xcolor}
\usepackage[sort,compress,super]{cite}
\usepackage{url}
\newcommand{\supprepo}{\url{https://github.com/MRDLab/femoral-sim-calibration}}

\makeatletter\renewcommand{\copyright}{{\ooalign{\hfil\raise.07ex\hbox{c}\hfil\crcr\mathhexbox20D}}}\makeatother
\makeatletter
\newcommand{\tblcap}[1]{\caption{#1}{\NINE\noindent\unhbox\captionbox\par}\vskip5pt}
\makeatother
\begin{document}

\catchline{0}{0}{2027}{}{}

\markboth{A. Manicka et al.}{Physics-Calibrated Simulation for Robotic Femoral Access}

% \title{Physics-Calibrated Simulation of Deformable Tissue for\\ Ultrasound-Guided Robotic Femoral Access}
\title{Simulating Vein Compressibility for Ultrasound-Guided Femoral Access: Speed, Accuracy, and Phantom Validation}

\author{Amit Manicka$^{a,*}$, Laurent Coulon, PhD,$^{b}$ Darwin Ang MD, PhD, MPH$^{b,bb}$ \\Michael Loushin, MD,$^{c}$ and Timothy M. Kowalewski, PhD$^{a,*}$}

\address{$^a$Department of Mechanical Engineering, University of Minnesota,\\ Minneapolis, Minnesota 55455, United States of America}

\address{$^b$Lightside Surgical, Bellevue, Washington 98005, United States of America}
\address{$^{bb}$Healtcare Corporation of America, HCA HealthONE Trauma Quality Improvement Program Univ.,  Ocala, Florida }

\address{$^c$Visible Heart Laboratory, Department of Anesthesiology, University of Minnesota Medical School,\\ Minneapolis, Minnesota 55455, United States of America}

\address{$^*$Corresponding author. E-mail: manic022@umn.edu, timk@umn.edu}

\twocolumn[{%
\maketitle

\begin{abstract}
In ultrasound-guided femoral access, probe pressure deforms the tissue and changes the vascular anatomy a robot observes. A simulator must reproduce these mechanics fast enough for robotic interaction, yet the computational effort needed for a clinically meaningful signal like vein collapse before artery is unclear. We test the feasibility of a physics-based simulation in NVIDIA's Isaac Sim for Healthcare, with Stable Neo-Hookean tissue solved by Vertex Block Descent on one GPU. A simulation of a low-fidelity cylindrical benchtop phantom with known geometry and mechanical properties pressed by a robot predicted its artery within 5 percentage points at every depth. It matched the vein within 3 points through 11~mm of a 25.4~mm press, and followed the collapse behavior of the low-fidelity physical phantom. Across element size, time step and solver iterations, wall clock per frame scaled linearly with solver work, but accuracy did not improve with more work. The chosen setting cut the solver's time per frame factor of 976, from 29,700~ms to 30.4~ms, within 1.03 points of the finest reference setting. The physics solver alone took 30.4 ms per frame, and with ultrasound and graphics rendering took to full stack took 56.4~ms per frame. Finally, seeding the digital simulation with realistic literature-based mechanical properties, the leg model compressed the vein far more than the artery at every setting, reproducing the expected clinical signal (vein collapse before artery) use to distinguish them.  In total, this suggests that the NVIDIA Isaac Sim platform used may be suitable to augment training for reinforcement learning or online simulation for autonomous femoral access with ultrasound. 
\end{abstract}

}]
\section{Introduction}

Simulation is now a standard environment for training and evaluating medical robots\cite{jiang2023robotic,elmekki2025rlreview,guo2026challenges}, but the mechanics of current robot-learning frameworks\cite{mittal2023orbit,nvidia2026isaac} remain strongest for rigid bodies\cite{aljalbout2026realitygap,zafra2025simulators}, and probe-control policies for robotic ultrasound are typically trained on static image volumes or rigid scenes\cite{jiang2023robotic,shen2023volumetric,luo2025multimodal,elmekki2025rlreview}. Deformable solvers exist, but their calibration against real tissue is rarely reported\cite{zafra2025simulators,ou2025cressim,liang2024realtosim}, so a simulator can look anatomically correct while behaving mechanically inaccurately.

This matters because probe pressure, needle insertion and vessel compression all change the anatomy the robot observes\cite{jiang2023robotic,qin2026review}, and the ultrasound image measures the deformed anatomy rather than a fixed model\cite{jiang2022atlas,suligoj2021robust}. The chain runs from a robot action to a contact force, to the deformed tissue, to the image that drives the next action (Fig.~S1). Sim-to-real transfer depends more on observations that stay consistent with probe motion than on single-frame realism\cite{qian2026trfdf}.

We target fully automated access to both the common femoral vein and artery for emergency medicine and resuscitation. In these settings, aortic balloon occlusion needs arterial access\cite{stannard2011reboa}, extracorporeal life support can need both vessels\cite{swol2018ecls}, and an absent pulse makes ultrasound replace palpation\cite{saugel2017cvc,brattain2021aiguide}. Probe pressure collapses the thin-walled vein while the artery holds its lumen\cite{guerrero2006dvt}, and clinical guidance uses this compressibility to tell them apart\cite{lamperti2012recs,vegas2025ase}. Although color Doppler can also separate them\cite{saugel2017cvc}, we do not assume it is available in emergency use, and thus rely on the B-mode image. Most robotic ultrasound systems regulate contact force to suppress probe-induced deformation\cite{bao2021constantforce,jiang2022atlas,beber2024impedance,si2025teleop}. Only one has used compression to classify the vessels\cite{raina2026aura}. Compression also has a cost: a compressed vein is a smaller target, so clinicians press lightly during puncture\cite{alvarado2026msd}. We envision a brief, deeper press that labels the vessels, then a lighter contact that keeps the vein open for the needle. However, both need a model that predicts vein and artery lumen area at each probe depth. In the Seldinger technique, the needle, dilator and catheter each compress the vein again, and every later step inherits the puncture site chosen under the probe\cite{seldinger1953}. This work therefore starts with probe contact.

A useful simulator must reproduce this behavior fast enough for a robot to interact with it, and for the thousands of repetitions that learning demands\cite{mittal2023orbit,elmekki2025rlreview}. We report cost as wall-clock milliseconds per frame, where a frame is 16.7~ms of simulated time, one image at the 60~Hz rate of an ultrasound display. Speed is not the only concern: a numerically accurate solution can still rest on material parameters that do not represent real tissue\cite{aljalbout2026realitygap,liang2024realtosim}. \textit{This work therefore asks how much computational effort is required to reproduce a clinical signal faithfully enough for a robot to interact with?} We test whether Isaac Sim is feasible for this task. We build an interactive Stable Neo-Hookean model\cite{smith2018stableneohookean} of the femoral region, solved on the GPU with Vertex Block Descent\cite{chen2024vbd}, whose deformation drives the rendered ultrasound image. We first compare a simulation of a simple benchtop phantom of known materials with robot presses on it. We then map how element size, time step and iterations trade accuracy against speed, first on a tissue slab, then on the leg. Finally, we give the leg model literature-based tissue properties and test whether it separates the femoral vein from the artery. Supplementary material (Sections~S1--S3, Figs.~S1--S7, Tables~S1--S9), videos and the simulation code are available online.\footnote{\supprepo}

\section{Methods}
\subsection{Units and names}
\label{sec:units}
\emph{Simulated time} passes inside the model while \emph{wall-clock time} is what the computer takes. Frames advance in time steps $\Delta t$, often called substeps elsewhere. Lumen area is a percent opening of the same lumen at rest: differences in it are percentage points (``points''). A depth is \emph{settled} when lumen area and displacement vary by under 2\% over 30 frames (500~ms of simulated time). The \emph{leg model} is the simulated anatomy. The \emph{benchtop phantom} is the physical gel phantom, and the \emph{simulated phantom} is its model. The \emph{reference setting} is the finest leg setting completed, and the \emph{chosen setting} is the fast setting selected for use.

\subsection{Deformable tissue simulation}
\label{sec:vbd_method}
The tissue is a tetrahedral finite-element mesh with the Stable Neo-Hookean constitutive model\cite{smith2018stableneohookean}. It is solved with Vertex Block Descent (VBD)\cite{chen2024vbd}, a solver from the computer graphics literature whose per-vertex updates parallelize well on a GPU (Fig.~S2). Contact enters as penalty energies in the same per-vertex minimization rather than as a separate constraint solve. Each tetrahedron has a deformation gradient $\mathbf{F}=\mathbf{D}_s\mathbf{D}_m^{-1}$, where $\mathbf{D}_s$ and $\mathbf{D}_m$ hold the current and rest edge vectors from vertex 0\cite{irving2006invertible}. Its strain energy density is
\begin{equation}
    \Psi = \frac{\hat{\mu}}{2}(I_C-3) + \frac{\hat{\lambda}}{2}(J-\alpha)^2,
\end{equation}
with $I_C=\|\mathbf{F}\|_F^2$, $J=\det\mathbf{F}$ the local volume ratio, $\alpha=1+\hat{\mu}/\hat{\lambda}$, $\hat{\mu}=\mu$ and $\hat{\lambda}=\lambda+\mu$, where the Lam\'e parameters $\mu=E/[2(1+\nu)]$ and $\lambda=E\nu/[(1+\nu)(1-2\nu)]$ follow from Young's modulus $E$ and Poisson ratio $\nu$ (Table~\ref{tab:anatomy}), so the energy matches linear elasticity at rest\cite{smith2018stableneohookean}. A tetrahedron straddling two materials takes the mean of its sub-samples, excluding bone. Inverted elements ($J<0$) keep a defined energy\cite{smith2018stableneohookean}, so the solver continues past them. At each time step VBD minimizes
\begin{equation}
    \mathcal{E}(\mathbf{x}) = \sum_i \frac{m_i}{2\Delta t^2}\left\|\mathbf{x}_i-\tilde{\mathbf{x}}_i\right\|^2 + \sum_e V_e\,\Psi_e(\mathbf{x}) + \mathcal{E}_{\mathrm{contact}}(\mathbf{x}),
\label{eq:vbd_obj}
\end{equation}
where $m_i$ is the mass of vertex $i$, $\tilde{\mathbf{x}}_i$ its inertial target, $V_e$ and $\Psi_e$ the rest volume and energy density of element $e$, and $\mathcal{E}_{\mathrm{contact}}$ the probe and vessel-wall contact energies (Section~\ref{sec:contact}). VBD moves one vertex at a time with the others fixed, by a Newton step on the $3\times3$ block Hessian of the elements and contacts touching that vertex. Vertices sharing no element are independent. A graph coloring groups them into colors, each color is updated in parallel, and one iteration is one pass over all colors. Iterations per step are a fixed count, with no error tolerance and no subdivision of the time step. The solver (Newton~1.2.1) adds a rate-dependent damping stress with $\gamma=1.0$~s in every element, which vanishes at rest (Section~S1).

In each frame, the deformed tissue surfaces are passed to raysim, the ultrasound simulator of NVIDIA's Isaac for Healthcare\cite{nvidia2026isaac}, which ray traces a B-mode image from them. Isaac Sim renders the probe and the leg in the same loop. All runs used one workstation with an NVIDIA GeForce RTX~5090 GPU (32~GB), an Intel Core Ultra~9~285 CPU (24 cores) and 64~GB of RAM under Ubuntu~24.04.5~LTS. The solver was Newton~1.2.1 on NVIDIA Warp~1.13.0, built against CUDA~12.9. Isaac Sim~6.0.1 rendered the probe, leg and room, and the Isaac for Healthcare ray tracer (\texttt{i4h-sensor-simulation}~0.4.0) ran in a separate container on CUDA~12.8\cite{nvidia2026isaac}.

\subsection{Probe contact and measurements}
\label{sec:contact}
Probe--tissue contact is a linear penalty spring between the rigid probe and a contact sphere of radius 0.25~$\times$ element size on each tissue node,
\begin{equation}
\mathcal{E}_{\mathrm{probe}} = \tfrac{1}{2}k_e d^2, \qquad \mathbf{f}_n = k_e d\,\mathbf{n},
\label{eq:probe}
\end{equation}
where $d$ is the sphere's penetration of the probe surface, $\mathbf{n}$ the surface normal. $k_e=2000$~N/m is a numerical penalty parameter tuned by hand so that the probe face did not visibly pass through the skin. A node of mass $m=\rho\,\Delta x^3$ on this spring oscillates with period $T=2\pi\sqrt{m/k_e}$, the contact spring's period: 0.258~ms for 1.5~mm elements and $\rho=1000$~kg/m$^3$, between the tabulated densities of fat and muscle\cite{itis2025database}. Opposite vessel walls closer than $r=0.4\times$ element size are held apart by a quadratic penalty and, below $\tau=r/2$, a logarithmic barrier. For two wall points $\delta$ apart, $\mathcal{E}_{\mathrm{wall}}=\tfrac{1}{2}k_e(r-\delta)^2$ for $\tau\le\delta<r$ and $\tfrac{1}{2}k_e\tau^2[1-2\ln(\delta/\tau)]$ for $\delta<\tau$, with energy and force continuous at $\tau$ (Section~S1). The contact energy in Eq.~(\ref{eq:vbd_obj}) is then
\begin{equation}
\mathcal{E}_{\mathrm{contact}}=\sum_{j}\mathcal{E}_{\mathrm{probe}}(d_j)+\sum_{(a,b)}\mathcal{E}_{\mathrm{wall}}(\delta_{ab}),
\label{eq:contact}
\end{equation}
over tissue nodes $j$ penetrating the probe and wall-point pairs $(a,b)$ closer than $r$. The walls never meet, so the mesh cannot represent a lumen area below a minimum, the \emph{closure floor} (Section~\ref{sec:vascular}).

The probe, a flat $40\times12$~mm face, was displacement-controlled. A sweep first rotated the drawn probe from $-45^\circ$ to $+45^\circ$ about the limb axis at $s=20$~mm, hovering 2~mm above the skin, so only the image changed (Fig.~S3). The press then advanced at $0^\circ$ by 0.4~mm per frame to 3, 7, 11, 15, 19, 23 and 25.4~mm, holding each depth until it settled. Lumen area is measured on the $s=20$~mm cut (Section~S1). Probe force can be derived from the contact term but is not reported (Section~\ref{sec:limitations}).

\subsection{Benchtop phantom: a low-fidelity feasibility test}
\label{sec:phantom_methods}
\begin{figure}[!t]
    \centering
    \includegraphics[width=0.85\columnwidth]{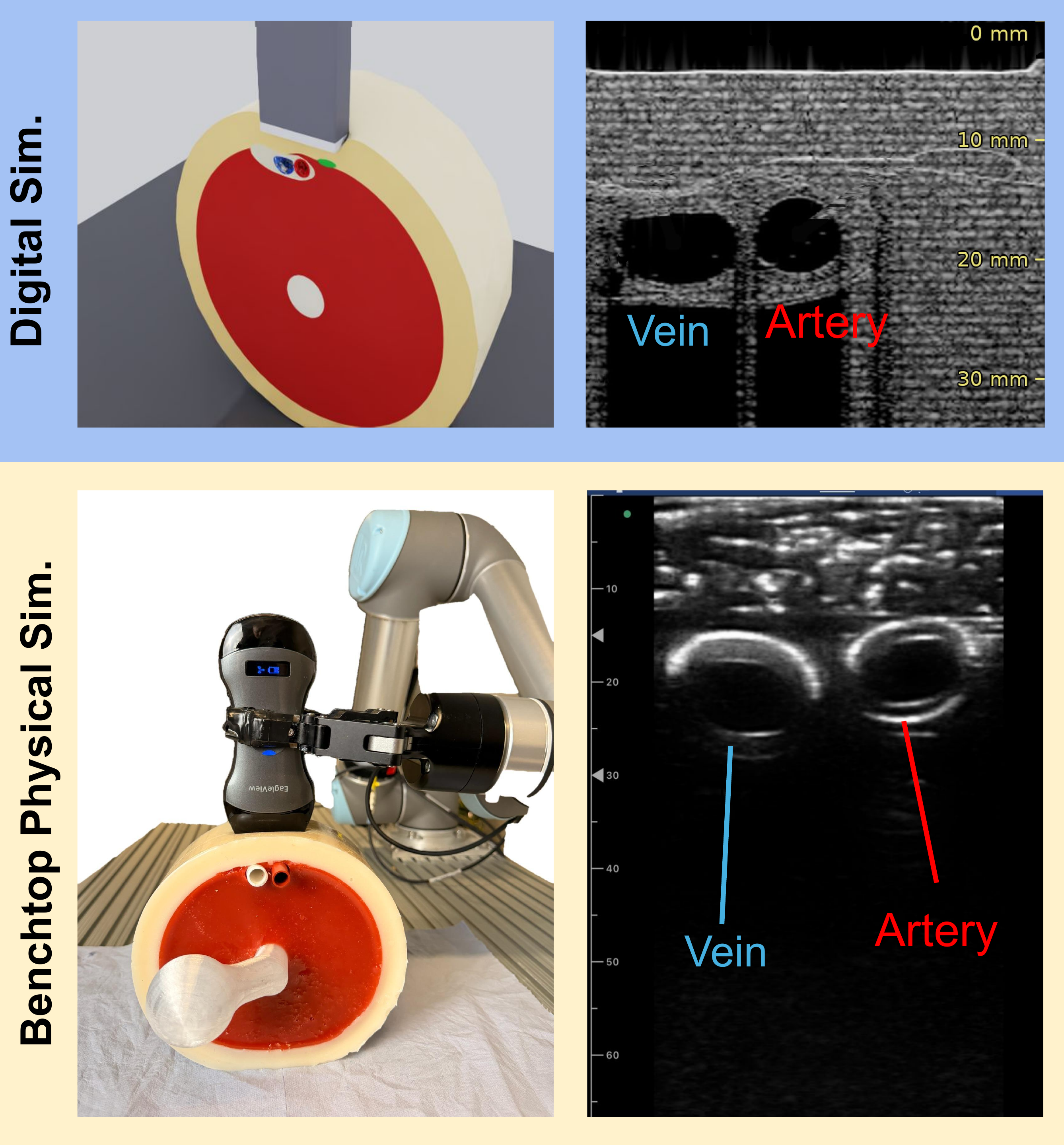}
    \caption{Phantom validation setup. Top: the simulated phantom, cut open at the ultrasound plane,
    and its ray-traced B-mode image. Bottom: the UR5 robot pressing the linear probe on the benchtop phantom, and the
    scanner's B-mode image. Vein (blue) and artery (red) are labeled in both images.}
    \label{fig:sim_vs_real}
\end{figure}
\begin{table*}[!t]\centering
\tblcap{Structures modeled, as meshed at $s=20$~mm. Depths are
perpendicular to the skin, except the CFV outer far wall. Its 29.2~mm is vertical, below the front of the thigh, and 28.2~mm perpendicular to the skin. (set): a dimension assigned for the model where no measurement applied, as is the 10.0 mm artery--vein center spacing. Each E and $\nu$ was chosen with reference to published measurements (Section~\ref{sec:limitations}). E is nominal, and elements at material interfaces take blended values.\label{tab:anatomy}}
{\footnotesize\setlength{\tabcolsep}{3pt}%
\begin{tabular}{@{}>{\raggedright\arraybackslash}p{2.4cm}>{\raggedright\arraybackslash}p{\dimexpr\linewidth-4.25cm-6\tabcolsep\relax}>{\centering\arraybackslash}p{1.15cm}>{\centering\arraybackslash}p{0.7cm}@{}}
\hline
\textbf{Structure} & \textbf{Geometry (source)} & \textbf{$E$ (kPa)} & \textbf{$\nu$} \\
\hline
Skin and fat (one body) & shell 16.0~mm; limb diameter 184.9~mm (set) & 3 & 0.45 \\
Muscle & core, radius 76.5~mm & 20 & 0.45 \\
CFA wall & lumen 7.0~mm, wall 1.0~mm (set); intima 19.5~mm deep\cite{zambetti2025diameter} & 600 & 0.45 \\
CFV wall & lumen 10.0~mm, wall 0.5~mm (set); intima 17.8~mm deep\cite{zambetti2025diameter}; outer far wall 29.2~mm vertical (28.2~mm perpendicular) & 20 & 0.45 \\
Femoral nerve & ellipse $9.6\times3.7$~mm\cite{lonchena2016femoral}; 13.0~mm lateral to CFA (about 12.7~mm\cite{gray16anatomy}; measured $11.1\pm2.9$~mm\cite{muhly2011crease}) & 40 & 0.45 \\
Femoral sheath & constant section in the mesh (funnel in CAD only); $s=0$--35~mm (set; usually 30--40~mm long\cite{moore2018coa}); wall thickness set & 20 & 0.45 \\
Perivascular tissue & fills the sheath around the vessels & 3 & 0.45 \\
Femur & 23.4~mm rod on the limb axis (set) & pinned & -- \\
\hline
\end{tabular}}
\end{table*}
% Set from tables/phantom_parameters.tex (generated by scripts/make_tables.py) in the JMRR table format.
\begin{table*}[!t]\centering
\tblcap{The benchtop phantom against the anatomical model. Leg model (CAD): the cross-section 20~mm below the inguinal
ligament, as meshed, with the benchtop phantom's two simplifications (no pubic ramus, 81~mm long), and moduli in Table~\ref{tab:anatomy}. Benchtop: the mold's design dimensions and catalog tube sizes, not caliper measurements of the benchtop phantom. Moduli are supplier values. It has no sheath, femoral canal or nerve.\label{tab:phantom_parameters}}
{\footnotesize\setlength{\tabcolsep}{3pt}%
\begin{tabular}{@{}>{\raggedright\arraybackslash}p{\dimexpr\linewidth-9.5cm-4\tabcolsep\relax}>{\centering\arraybackslash}p{4.0cm}>{\centering\arraybackslash}p{5.5cm}@{}}
\hline
\textbf{Parameter} & \textbf{Leg model (CAD)} & \textbf{Benchtop} \\
\hline
Length along the limb & 81~mm & 81~mm \\
Diameter & 184.9~mm & 184.1~mm \\
Skin and fat thickness & 16.0~mm & 15.9~mm \\
Artery lumen diameter & 6.96~mm & 9.00~mm \\
Vein lumen diameter & 9.95~mm & 10.00~mm \\
Artery wall thickness & 1.00~mm & 1.00~mm \\
Vein wall thickness & 0.50~mm & 1.50~mm \\
Skin to artery inner wall & 19.5~mm & 22.6~mm \\
Skin to vein inner wall & 17.8~mm & 20.8~mm \\
Artery to vein, center to center & 10.0~mm & 16.0~mm \\
\hline
Tissue model & Neo-Hookean; moduli in Table~\ref{tab:anatomy} & Neo-Hookean; values below \\
Soft-tissue material & -- & SimuGel blends (muscle \#0+\#1, fat \#2+\#0) \\
Young's modulus, muscle / fat & -- & 510 / 380~kPa (supplier, mixed by mass) \\
Vessel walls & -- & silicone tubes, 70A artery / 35A vein \\
\hline
\end{tabular}}
\end{table*}
The simulation was first tested on a benchtop phantom molded from the leg design (Section~\ref{sec:anatomy}, Table~\ref{tab:phantom_parameters}). The phantom is deliberately low in fidelity, but its materials have supplier-specified properties. It has a muscle core of radius 76.2~mm inside a 15.9~mm fat layer, 81~mm long, both blends of a synthetic tissue gel\cite{humimic2026simugel} whose Young's moduli, 510 and 380~kPa, are the supplier's values mixed by mass. Silicone tubes stand in for the vessels: the artery tube has a 9~mm bore and a 1.0~mm wall of Durometer 70A, the vein tube a 10~mm bore and a 1.5~mm wall of 35A. Their moduli, 5,500 and 1,400~kPa, follow from durometer by Gent's relation\cite{gent1958hardness}. The tubes are water-filled and open at both ends, so no pressure builds. The femur is a printed copy held only by the gel, and there is no sheath, femoral canal or nerve. Its model, the simulated phantom (Fig.~\ref{fig:sim_vs_real}), used the leg's element sizes and solver settings (Section~\ref{sec:calib_methods}), $\nu=0.49$ for gels and tubes, as for nearly incompressible solids\cite{mott2008bulk,fougeron2024poisson}, a free femur at 200,000~kPa, and for walls thinner than one element the modulus that keeps their bending stiffness, $E_\mathrm{eff}=E\,(t/1.5~\mathrm{mm})^3$ (Table~S6). No input was adjusted to the phantom.

A UR5 robot (Universal Robots, Odense, Denmark) held a linear ultrasound probe on the phantom 20~mm below the ligament's level. Zero was the first probe height that gave a visible image. The probe was pressed to the seven depths three times each, withdrawn between depths, and held 5~s before each image. Inner lumen height was measured between the inner wall echoes at one column per vessel: a Python script (NumPy, SciPy) found the echo peaks and the operator chose the inner pair (script, picks and every image in the supplement, Fig.~S7). Height was converted with the scanner's 22.24~pixels/mm and the 1480/1540~m/s ratio of sound speeds in water and in the scanner's model, then to area through a rounded-rectangle model, $A/A_0=x(2-x)$ with $x=h/D_0$ the inner height over the catalog bore. Heights at first contact were 8 to 10\% shorter and served as a sensitivity check. Agreement at a press depth was defined before the presses to be convergence of the simulation and benchop within 5 percentage points. Close agreement is defined within 3 percentage points, which on the same basis as the numerical tolerance\cite{ogata2010modern}.

\subsection{Leg model: anatomy and materials}
\label{sec:anatomy}
The anatomy is a parametric CAD model of the right femoral region, dimensioned from the literature (Table~\ref{tab:anatomy}). The segment runs from the inguinal ligament to $s=81$~mm, the length of the benchtop phantom. Its reference level, $s=20$~mm, lies inside the arterial window that ends at the profunda femoris origin\cite{moore2018coa}. It is a model level, not a recommended puncture site: a safe arterial puncture lies on the common femoral artery below the inguinal ligament and above its bifurcation, located on each patient's ultrasound\cite{moore2018coa,seto2010faust}. The mesh extrudes the $s=20$~mm cross-section along the limb, so each body has a constant section (Table~\ref{tab:anatomy}). Vessel lumen targets are the duplex ultrasound means of Zambetti \textit{et al.} ($n=983$)\cite{zambetti2025diameter}, measured at the saphenofemoral junction, a different level from ours. Artery diameter also varies along the vessel\cite{pruna2019cfa} and with age, sex and body size\cite{sandgren1999cfa}. As built, both lumens are smaller than their targets because the CAD read each target lumen diameter as the outer wall diameter: the common femoral artery (CFA) lumen is 7.0~mm against $9.0\pm2.1$~mm and the common femoral vein (CFV) lumen 10.0~mm against $11.0\pm2.7$~mm\cite{zambetti2025diameter}. Skin-to-intima depths, 19.5~mm (CFA) and 17.8~mm (CFV), lie within 0.35~SD of the cohort means\cite{zambetti2025diameter}. The femur is the only bone and its nodes are pinned. Every modulus and Poisson ratio is an informed modeling choice made with reference to published measurements of each tissue\cite{segain2024,chakouch2015mre,bensamoun2006mre,alkhouli2013adipose,samani2007breast,feng2022,jadidi2019,schulzebauer2002,lebahn2025,wesly1975vein,vesely2015saphenous,song2022,sanchezramirez2026nerve,otsuka2018,mott2008bulk} (Section~\ref{sec:limitations}). The vessels are treated as voids with no internal pressure.

% Set from tables/simulation_geometry.tex (generated by scripts/make_tables.py) in the JMRR table format.
\begin{table*}[!t]\centering
\tblcap{Geometry and numerical settings of the slab and leg meshes. Sizes are the bounding box
of the resting mesh, with lateral across the limb, axial along it and depth from the skin inward.
The leg column gives the chosen setting, and the reference setting uses a 0.099~ms time step
(168 time steps per frame) with 140 iterations per step. (set): assigned for the model.
The damping coefficient $\gamma$ is in seconds, as in the solver's damping form (Section~S1).\label{tab:simulation_geometry}}
{\footnotesize\setlength{\tabcolsep}{3pt}%
\begin{tabular}{@{}>{\raggedright\arraybackslash}p{2.35cm}>{\raggedright\arraybackslash}p{\dimexpr(\linewidth-2.35cm-4\tabcolsep)/2\relax}>{\raggedright\arraybackslash}p{\dimexpr(\linewidth-2.35cm-4\tabcolsep)/2\relax}@{}}
\hline
\textbf{Parameter} & \textbf{Slab (parameter study)} & \textbf{Leg model} \\
\hline
Domain & block cut from the limb 20~mm below the inguinal ligament; side and deep faces held fixed & whole thigh segment, 81~mm long, held by a small fixed patch where the thigh rests on the table \\
Lateral width & 72.0~mm & 184.9~mm \\
Axial length & 16.5~mm & 81.0~mm \\
Depth & 82.5~mm & 184.9~mm \\
Nodes & 31,740 & 23,688 \\
Tetrahedra & 166,386 & 129,783 \\
Nodes held fixed & 1,896 (side and deep faces, bone) & 352 (bone 342, table patch 10) \\
Element size & 1.5~mm, uniform & 1.5~mm near the vessels; 5~mm femur; 30~mm elsewhere \\
Time step & 0.0186~ms (896 per frame) & 8.33~ms (2 per frame) \\
Iterations per time step & 140 & 35 \\
Contact stiffness $k_e$ & 2,000~N/m & 2,000~N/m \\
Damping $\gamma$ (set) & 1.0~s & 1.0~s \\
\hline
\end{tabular}}
\end{table*}

\subsection{Solver calibration}
\label{sec:calib_methods}
Numerical parameters were first studied on a slab cut from the leg at $s=20$~mm (Table~\ref{tab:simulation_geometry}), then on the leg. The slab is a rectangular cut of the leg's cross-section on a regular grid, with the artery, vein and nerve at the same positions within 0.55~mm (Fig.~S3). Element sizes of 2.5, 2.0, 1.5, 1.25 and 1.0~mm were run at a 0.0186~ms time step with 140 iterations per step. Since settling shows only that a run has stopped moving, mesh dependence was judged by comparing vein closure between sizes. Accuracy was judged against a reference run on each mesh: on the slab the 0.0186~ms, 140-iteration run. On the leg, it is a 0.099~ms time step with 140 iterations, the finest leg setting that completed without GPU failure. Both are numerical references, not exact solutions, and both still change with element size. A setting passes when vein and artery lumen areas stay within 3 points of the reference at every depth. This tolerance was set before the repeat runs and follows the 2 to 5\% bands that define settling time in controls theory \cite{ogata2010modern}. The leg mesh is graded, with 1.5~mm elements near the vessels and coarser ones elsewhere (Table~\ref{tab:simulation_geometry}). Each solver iteration is a fixed series of GPU kernel launches, which the CPU otherwise issues one by one. A CUDA Graph records the series once, and the CPU then submits it with a single call\cite{nvidia2026cuda}.

\section{Results}
\subsection{Benchtop phantom}
\label{sec:phantom_results}
\begin{figure}[!t]
    \centering
    \includegraphics[width=\columnwidth]{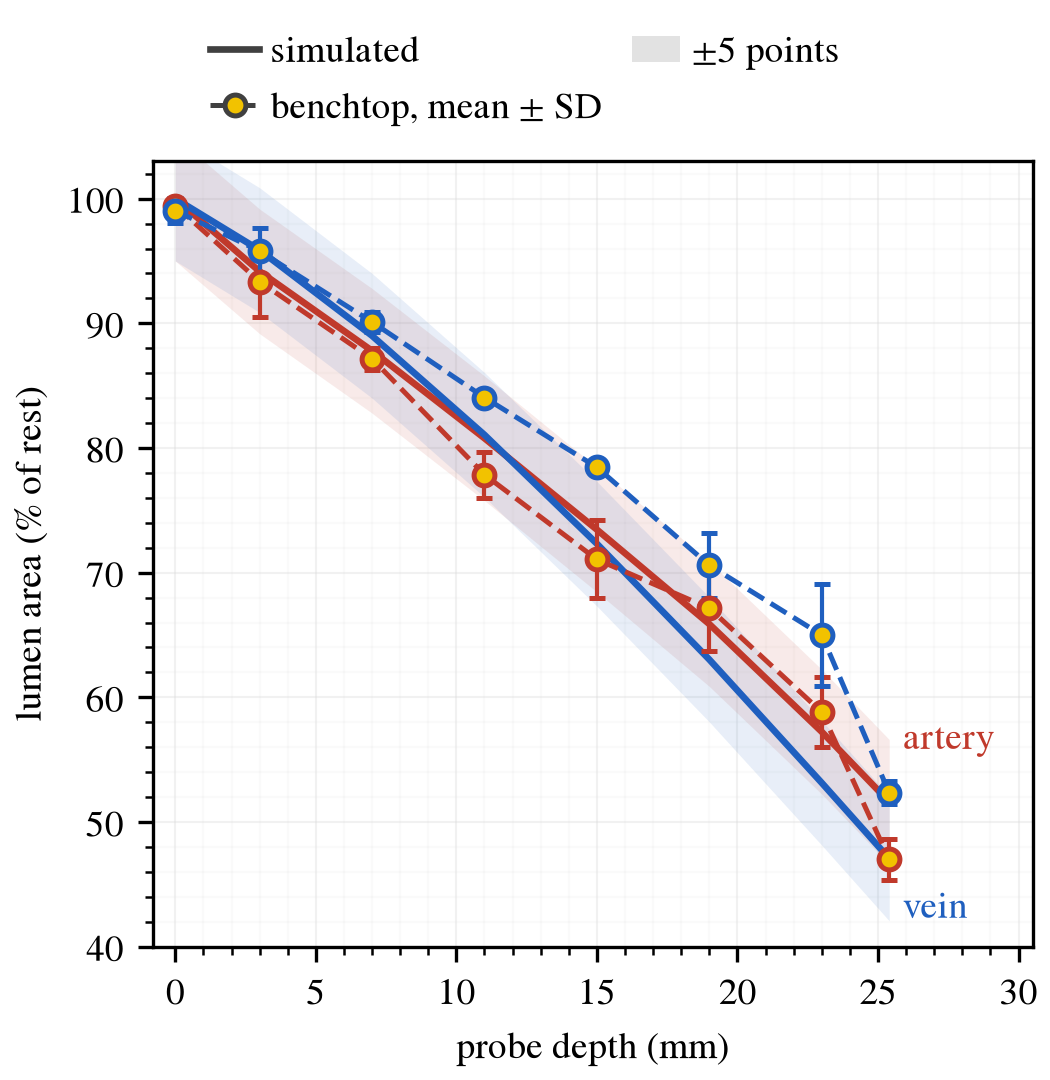}
    \caption{Benchtop phantom against the digitally-simulated low-fidelity phantom: lumen area against probe depth, at 20~mm below the inguinal ligament. Lines: simulated phantom, reference setting. Markers: benchtop phantom, mean $\pm$ SD of three robot presses (one 25.4~mm press from a separate session). Bands: $\pm5$ $\%$ about each simulated line, the agreement tolerance set before the presses. }
    \label{fig:cast_validation}
\end{figure}
Fig.~\ref{fig:cast_validation} compares the two phantoms (Table~S5, with every image in Fig.~S7). The artery agreed within a 5 $\%$ error at all seven depths, and within 3 at six of them. The vein agreed within 3 $\%$ through 11~mm. From 15 to 25.4~mm, the simulated phantom collapsed the vein more than the benchtop phantom did, by 5.3 to 11.9 $\%$, always with the same sign. With the first-contact heights as rest diameters, the artery still agreed within 5 $\%$ at every depth while the vein discrepancies grew to at most 16.4 $\%$. Neither phantom separated the vessels. At every depth their vein and artery lumens stayed within 8 $\%$ of each other on the benchtop and 5 in simulation. On this material, about 25 times stiffer than soft tissue, the chosen setting differed from the reference by up to 4.4 $\%$, so the comparison uses the reference setting.

\subsection{Slab: element size, time step and iterations}
\label{sec:results_slab}
The slab set the element size, time step and iteration count carried to the leg (Section~S2). At a 0.0186~ms time step, vein closure at 25.4~mm still changed with element size, from 33.8\% of rest at 1.5~mm to 19.5\% at 1.0~mm (Fig.~S4, Table~S1). The 1.5~mm size was chosen on cost, although a finer mesh still changed the result. Coarsening the time step 448-fold, to 8.33~ms, moved vein closure by $1.89\pm0.02$ $\%$ over three runs (Fig.~S5, Table~S2). Two settings failed to settle at some depths: their time steps, 0.249 and 0.980~ms, are 0.96 and 3.80 times the contact spring's 0.258~ms period. At 2.08~ms, 35 iterations stayed within $0.39\pm0.05$ $\%$ of the 140-iteration reference, so the leg model used 35 iterations per step.

\subsection{Leg: speed and accuracy}
\label{sec:results_leg}
\begin{figure}[!t]
    \centering
    \includegraphics[width=\columnwidth]{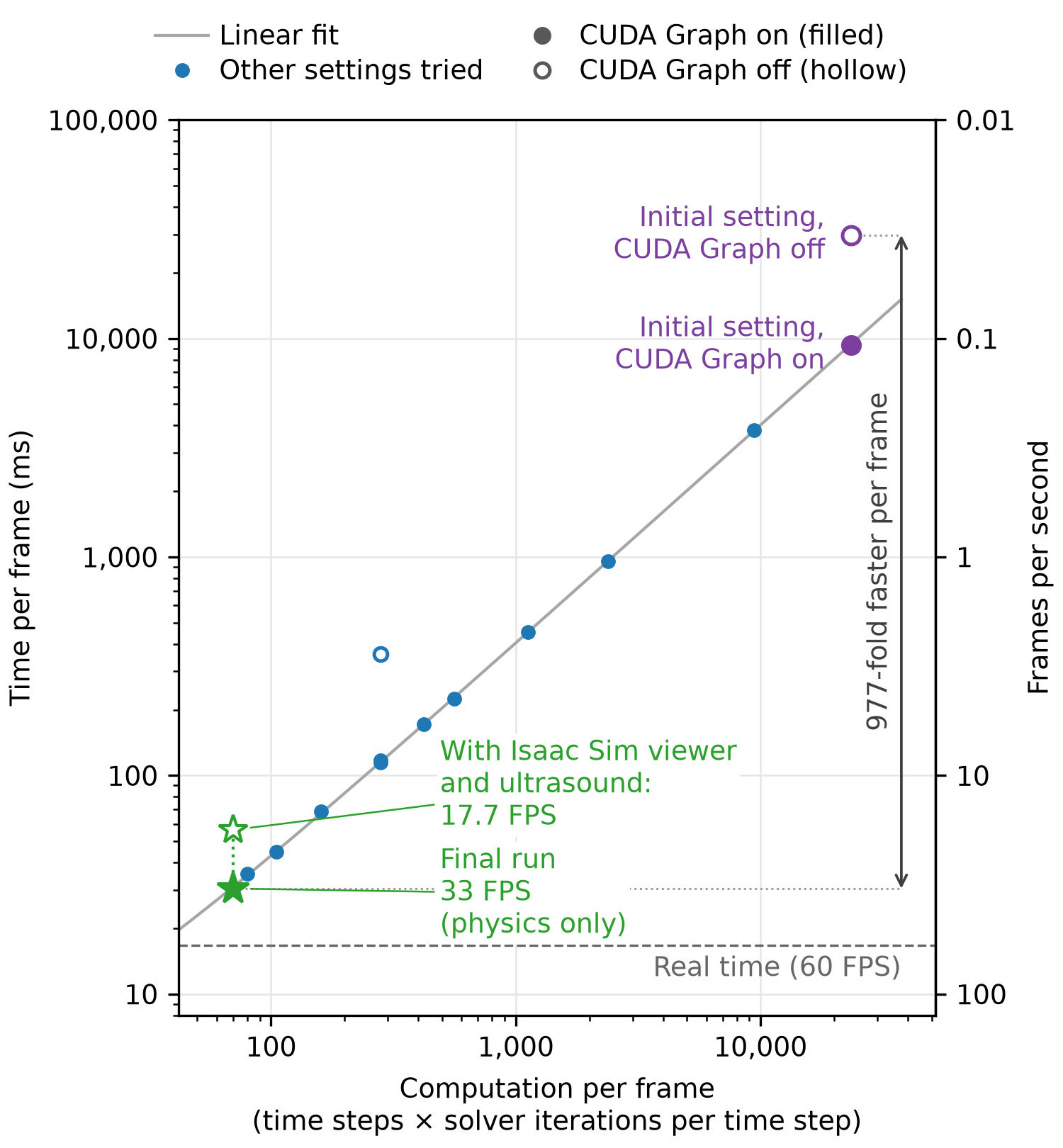}
    \caption{Leg model on one RTX 5090 with no other GPU load: wall clock per simulated frame against solver work per frame (time steps per frame $\times$ iterations per step), with the equivalent frames per second on the right. Filled circles: the twelve solver settings with CUDA Graphs on, mean of five runs (SD 0.4--2.3\% of the mean, smaller than the marker, Table~S4). Two settings share work 280. Open circles: CUDA Graphs off (Table~S3). ``Initial setting'': the reference setting. Filled star, ``final run'': the chosen setting, solver only (30.4~ms per frame). Open star: the chosen setting with the 3-D view and ray-traced ultrasound on the same GPU (56.4~ms, one run). Dashed: the 16.7~ms real-time limit. Line: two-term fit to the sixty CUDA-Graphs-on runs, 0.401~ms per solver pass plus 2.82~ms per frame (Fig.~S6, Table~S3).}
    \label{fig:speedup}
\end{figure}
CUDA Graphs removed the host's per-launch cost, a threefold gain at equal settings (Table~S3), and moved closure at 25.4~mm by under 1~point. Fig.~\ref{fig:speedup} and Table~S4 give every leg setting. Wall clock followed solver work at $0.400\pm0.001$~ms per solver pass (Table~S3). Accuracy did not follow the work increase. All twelve settings stayed inside the 3-point tolerance, and the largest gap, 1.20 $\%$, belonged to the 2.08~ms step with 35 iterations, which did four times the chosen setting's work. The chosen setting, solver only, took 30.4~ms per frame, 977 times faster than the reference without CUDA Graphs. Over five runs it stayed within $0.76\pm0.09$ (vein) and $1.03\pm0.00$ (artery) $\%$ of the reference mean at every depth, while repeat runs of the reference differed from their mean by $0.23\pm0.16$ and $0.13\pm0.06$ $\%$. The full loop, adding ray-traced ultrasound and the rendered 3-D scene, took 56.4~ms per frame in one run and stayed within 0.60 (vein) and 1.22 (artery) points of the reference mean. Every leg setting settled at every depth in all five runs.

\subsection{Predicted vessel response in the leg model}
\label{sec:vascular}
\begin{figure}[!t]
    \centering
    \includegraphics[width=\columnwidth]{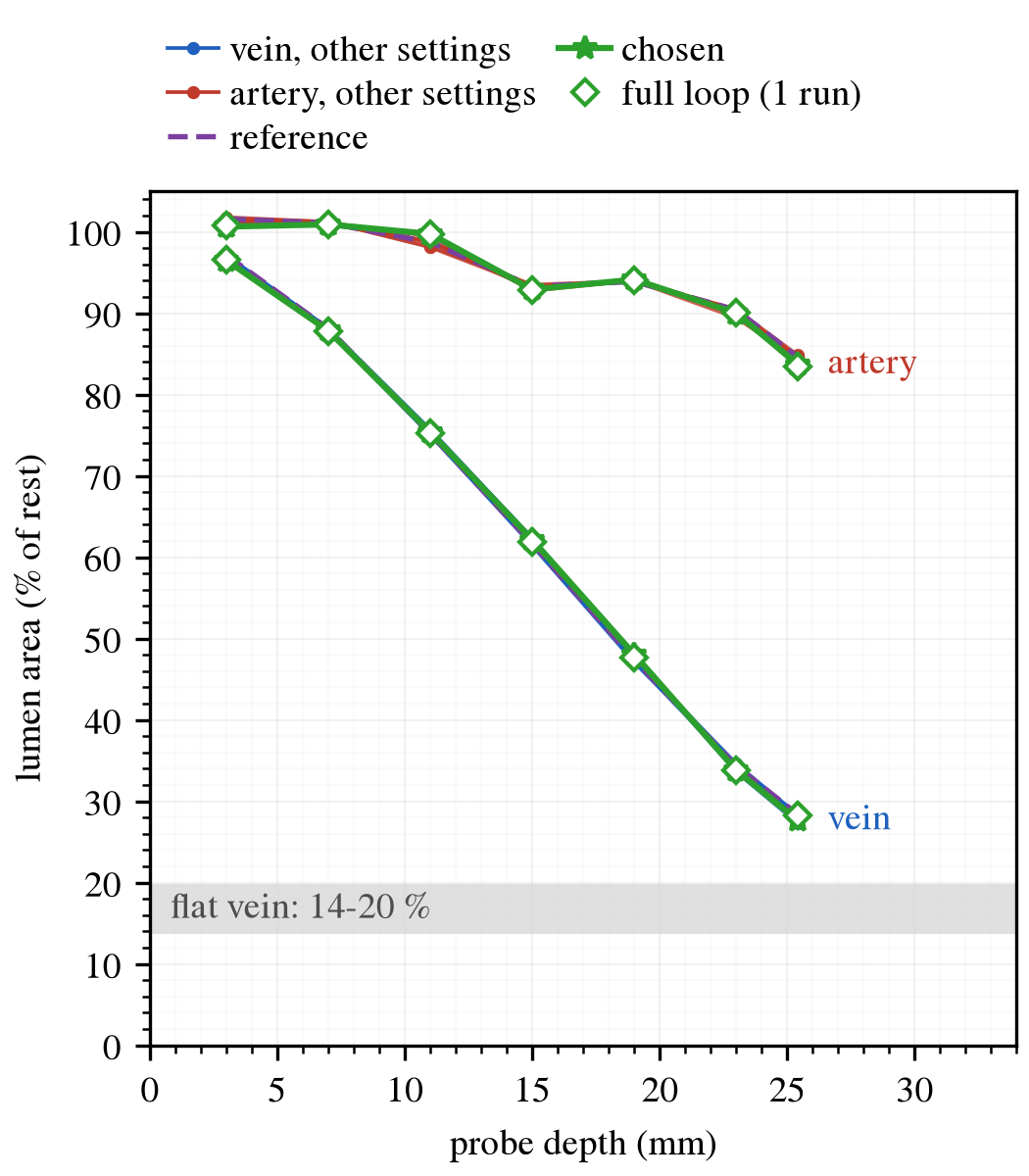}
    \caption{Leg model in Isaac Sim seeded with realistic mechanical properties for muscle and fat: vein (blue) and artery (red) lumen area against probe depth, for all twelve solver settings searched. Thin lines: the ten intermediate settings. Purple dashed: the reference setting. Green stars: the chosen setting ($\pm$ SD). Each is the mean of five runs. Hollow green diamonds: one run of the full loop. Gray band: the closure floor, the nominal reading of a fully flattened vein on this mesh, derived from contact spacing. }
    \label{fig:vein_artery}
\end{figure}
The vein compressed more than the artery at every depth and every solver setting (Fig.~\ref{fig:vein_artery}). At 11~mm the vein kept 75.4\% of its resting area and the artery 99.7\% (five runs). At 25.4~mm they kept $27.7\pm0.1\%$ and $83.7\pm0.1\%$, 56 points apart, where neither phantom's vessels differed by more than 8. At 11~mm the vessels had already separated by 24 points while the vein kept three quarters of its area as a target for the needle. How much separation a robot needs is not established here. The curve is indexed by probe depth without force, in vessels that hold no blood pressure, so it cannot predict the depth that suits a given patient. The vein reading stayed above zero because contact holds opposite walls about 1.2~mm apart ($0.8\times$ element size). This sets a nominal closure floor of 14--20\% on this mesh, and the reading stayed above it. On the leg, at most 859 elements (0.67\% of the mesh) were inverted at 25.4~mm, none in a vessel wall (Section~S2).

\section{Discussion}
\subsection{Agreement with the benchtop phantom}
The benchtop test asked whether the simulation could reproduce a simple object of known materials without tuning. It reproduced the artery at every depth and the vein through 11~mm. From 15~mm on the simulated vein always collapsed more than the benchtop vein. The discrepancy most likely comes from the vein tube's wall stiffness relative to the gel (Section~S3).

\subsection{From phantom to clinical parameters}
Neither phantom separated the vessels, so this test checked the solver, not the clinical signal. Once the solver had passed it, we gave the leg model literature-based tissue properties rather than building a more faithful phantom. The leg then compressed the vein far more than the artery at every solver setting (Section~\ref{sec:vascular}). The clinical signal came from the clinical inputs, not from the solver's cost. Whether the leg model carries the phantom's vein bias is untested, because its materials differ. Its predicted vein collapse at depth is a property of the model, not yet of a patient.

\subsection{Speed and accuracy trade-off}
On the leg, an 84-fold coarser time step and a quarter of the iterations kept closure well inside the tolerance set before the runs, above the reference's run-to-run spread. This is agreement with a numerical reference on one mesh, not with real tissue. On the slab, element size mattered more than any settled time step (Section~\ref{sec:results_slab}). We therefore suggest choosing the cheapest setting that settles at every depth and stays within tolerance of the reference, repeating both checks for every new mesh, material or probe motion.  At 56.4~ms per frame, the full loop is slower than the 16.7~ms real-time limit but within the 43 to 100~ms between actions of diffusion-policy controllers for robotic ultrasound\cite{qian2025latency}. The reference, at 29,700~ms per frame, falls far outside that range.

\subsection{Limitations}
\label{sec:limitations}
The vessels hold no blood pressure, flow, nor pulsation, which under ultrasound marks the artery, though a nearby vein can appear to pulse with it. In canines, venous pressure widens the femoral vein\cite{morris1974femoralvein}, rounding its section\cite{gero1968venous}. Arterial pressure would resist compression far more, so we expect pressure to widen the difference between the vessels. This omission is most serious in the emergencies we target. In shock, vein size changes with fluid status. In cardiac arrest the artery loses its pulse, and below a systolic pressure of 60~mmHg it can also compress under the probe\cite{saugel2017cvc}. Performance in these emergency states remains unknown.

No published stiffness was measured under this loading (Section~S3, Table~S6). Compared by shear modulus ($\nu=0.45$), our fat (1.0~kPa) is about twice the in vivo thigh median and our muscle (6.9~kPa) about quadruple\cite{segain2024}. The vein wall departs furthest from published data. Tensile tests give 5,700~kPa\cite{lebahn2025}, about 285 times our 20~kPa, which we chose to represent a thin wall's collapse under low transmural pressure\cite{gero1968venous}. Bending stiffness scales with $Et^3$, and our one-element wall is 1.5~mm thick against 0.33--0.35~mm in vivo\cite{alibazoner2021}. Walls matching the bending stiffness of a 0.34 or 0.50~mm wall (66.4 or 211.1~kPa) kept 35.3\% and 51.4\% of the vein's area at 25.4~mm against 27.6\% for 20~kPa, with the artery at 83.6--86.2\% in all three (Table~S6). Thus, the vein closes later but still compresses far more than the artery.

We do not report probe force. The endpoint is read from the image, the force error grows as elements invert (Section S2), and force was never calibrated against a known load. The benchtop comparison rests on one phantom at one level, with unmeasured supplier and durometer moduli and areas inferred from one inner height. Gel and silicone agreement does not show that the model predicts human tissue. The acoustic materials are NVIDIA's seven liver-workflow defaults\cite{nvidia2026isaac}, and the simulated image has not been compared with clinical images.

\section{Conclusion}
A simulation in Isaac Sim predicted a low-fidelity benchtop phantom's artery within 5 points at every depth. It matched the vein within 3 points through 11~mm, then collapsed the vein more than the phantom did. Robotic interaction does not require the finest solver setting, only an error small enough for the task. The solver's time per frame fell from 29,700~ms to 30.4~ms while the chosen setting stayed within 1.03 points of the reference. Because accuracy did not improve with more work, each candidate setting must be checked against a reference. Given literature-based tissue properties, the leg model kept the differential vein and artery deformation the ultrasound observation needs at every setting. Future work will add blood pressure and flow, validated force measurement and acoustic validation, then a physical robot in the loop on the benchtop phantom. With needle insertion added, success will mean first-attempt entry into the intended vessel\cite{seto2010faust}, no back-wall puncture and a guidewire inside the lumen\cite{saugel2017cvc}.

\nonumsection{Conflict of Interest}
\noindent L. Coulon and D. Ang are with Lightside Surgical, which intends to use this research in its product development. The other authors declare no conflict of interest.

\nonumsection{Acknowledgments}
\noindent The authors thank Professor Karthik Desingh and the Robotics: Perception and Manipulation Laboratory for computing resources. This research received no external funding.

\bibliographystyle{ws-jmrr}
\bibliography{references}

% photos/bios.tex: author photographs and biographies for the JMRR version (main_V7a.tex), set after the references
% as in the ws-jmrr template. Photos go in this folder under the names below (1 in wide, 300 dpi, black and white
% for print). Until a photo is added, a gray box with the file name stands in for it.
% Bio sources (2026-11-08): Lightside Surgical team page; UMN Mechanical Engineering profile (Kowalewski);
% UMN Institute for Engineering in Medicine news profile (Loushin). Each author should confirm their own text.
\newcommand{\authorphoto}[1]{\IfFileExists{photos/#1}{\includegraphics[width=1in]{photos/#1}}%
  {\fbox{\parbox[c][1.25in][c]{0.9in}{\centering\tiny photo\\ \texttt{#1}}}}}

\vspace{12pt}
\noindent\authorphoto{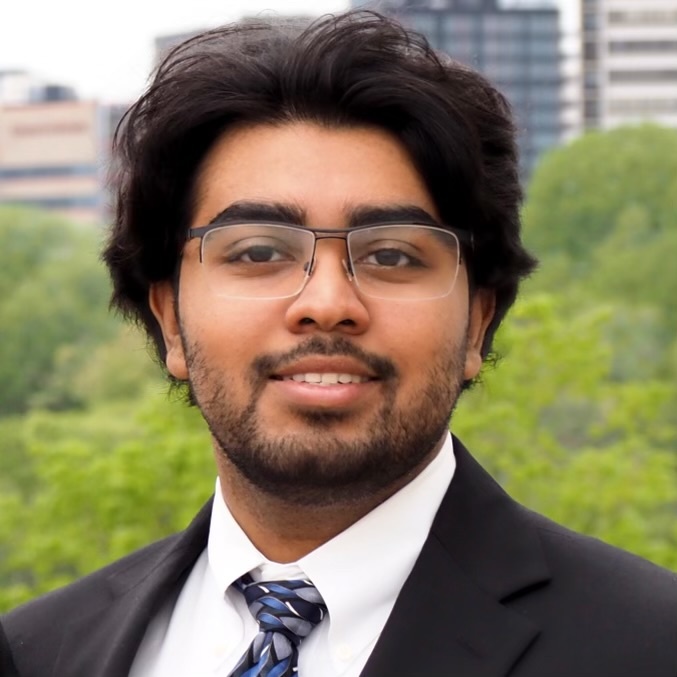}\par\vspace{4pt}

\noindent{\bf Amit Manicka} received his B.S. degrees in Mechanical Engineering and Computer Science from the University of Minnesota in 2024 and his M.S. degree in Mechanical Engineering from the University of Minnesota in 2025, where he worked on robotic impedance control. He is currently a Ph.D. student in the Department of Mechanical Engineering at the University of Minnesota while also pursuing his M.S. in Electrical Engineering concurrently. His research interests include robotic impedance control for force- and displacement-controlled insertions as it pertains to ultrasound-guided robotic vascular access.\\

\noindent\authorphoto{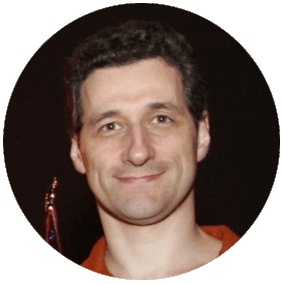}\par\vspace{4pt}

\noindent{\bf Laurent Coulon} graduated with his Ph.D. from \'Ecole Centrale in France. He is Co-Founder and Chief Software Officer of Lightside Surgical. He has over 25 years of experience building real-time systems for the gaming, film and medical industries, and has guided the development of more than 20 commercially shipped products. His technical work includes rigid and soft body dynamics, computational fluid dynamics, the finite element method and GPU acceleration. His current focus is simulating human tissue mechanics.\\

\noindent\authorphoto{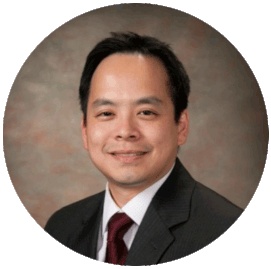}\par\vspace{4pt}

\noindent{\bf Darwin Ang}, M.D., Ph.D., M.P.H., FACS, is a trauma and critical-care surgeon and Co-Founder and Chief Medical Officer of Lightside Surgical. He served as enterprise trauma medical director for HCA and founded the UCF/HCA general surgery residency, where he built a robotic training pathway. He has authored more than 110 peer-reviewed publications and received two gubernatorial appointments to Florida's Trauma System Advisory Council. At Lightside Surgical, he helps translate robotic technology into surgical workflows.\\

\noindent\authorphoto{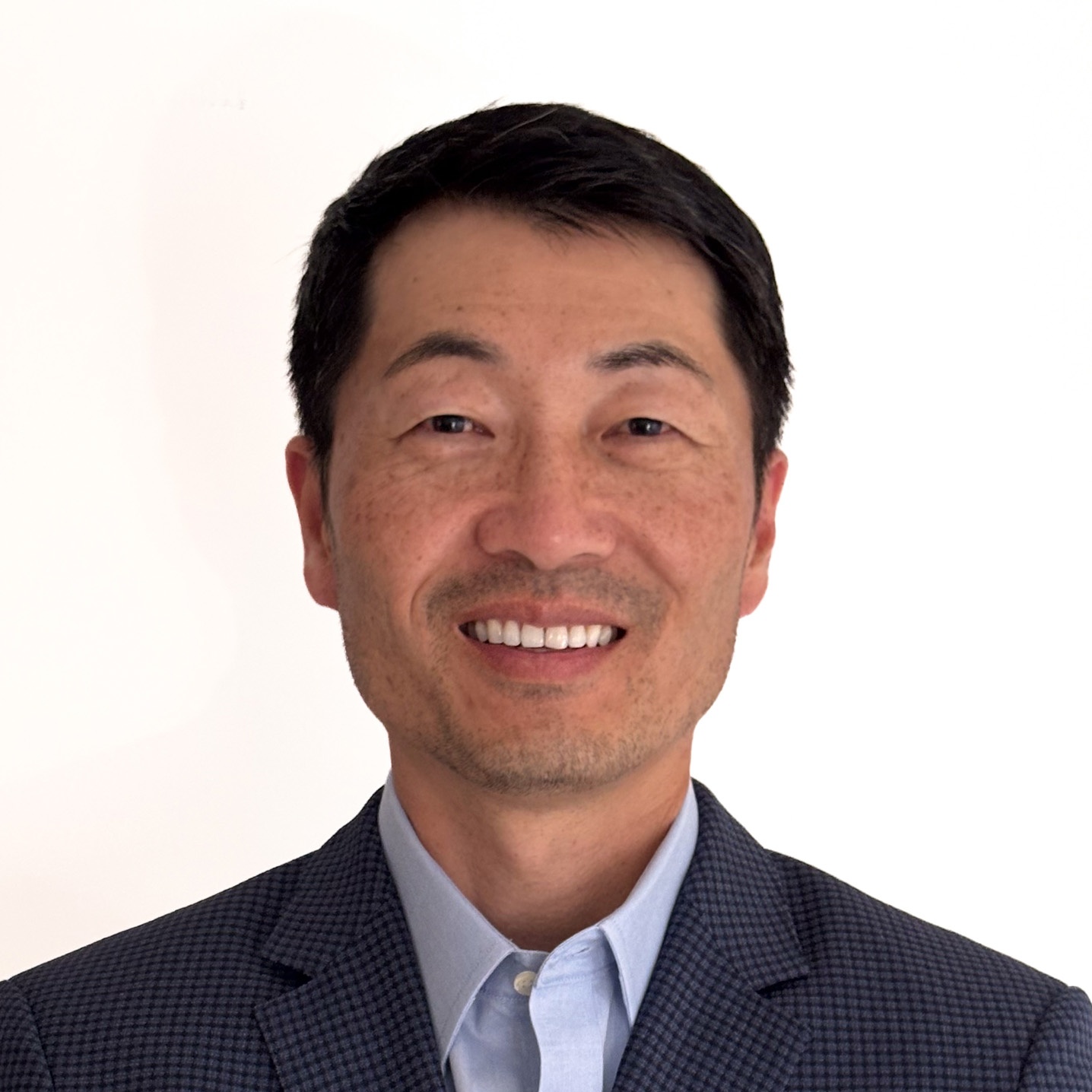}\par\vspace{4pt}

\noindent{\bf Michael Loushin} received his M.D. degree from the University of Minnesota Medical School. He was clinical faculty at the University of Minnesota from 2002 to 2016 through Twin Cities Anesthesia Associates, then practiced at Regions Hospital and Lakeview Hospital. He is a Professor in the Department of Anesthesiology and Associate Director of the Visible Heart Laboratories at the University of Minnesota, where he has worked since 1992. His research interests include translational systems physiology, thermoregulation, muscle physiology and medical technology.\\

\noindent\authorphoto{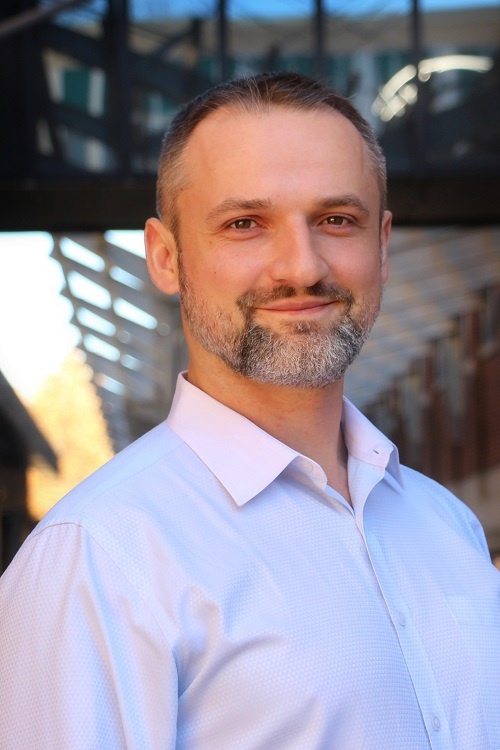}\par\vspace{4pt}

\noindent{\bf Timothy M. Kowalewski} received his B.S., M.S. and Ph.D. degrees in Electrical Engineering from the University of Washington, Seattle, in 2003, 2009 and 2012, respectively. He was the Richard and Barbara Nelson Assistant Professor of Mechanical Engineering at the University of Minnesota from 2012 to 2020, and has been an Associate Professor there since 2020. He is affiliated with the Minnesota Robotics Institute. His research interests include instrumented surgical tools, surgical robotics, and surgical simulation with virtual reality and realistic mock tissues.

\end{document}